# Accelerated materials language processing enabled by GPT

*Jaewoong Choi[1], Byungju Lee[1,*]*

[1]Computational Science Research Center, Korea Institute of Science and Technology, Seoul, Republic of Korea

[*]Corresponding author: Dr. Byungju Lee (blee89@kist.re.kr)

# Abstract

Materials language processing (MLP) is one of the key facilitators of materials science research, as it enables the extraction of structured information from massive materials science literature. Prior works suggested high-performance MLP models for text classification, named entity recognition (NER), and extractive question answering (QA), which require complex model architecture, exhaustive fine-tuning and a large number of human-labelled datasets. In this study, we develop generative pretrained transformer (GPT)-enabled pipelines where the complex architectures of prior MLP models are replaced with strategic designs of prompt engineering. First, we develop a GPT-enabled document classification method for screening relevant documents, achieving comparable accuracy and reliability compared to prior models, with only small dataset. Secondly, for NER task, we design an entity-centric prompts, and learning few-shot of them improved the performance on most of entities in three open datasets. Finally, we develop an GPT-enabled extractive QA model, which provides improved performance and shows the possibility of automatically correcting annotations. While our findings confirm the potential of GPT-enabled MLP models as well as their value in terms of reliability and practicability, our scientific methods and systematic approach are applicable to any materials science domain to accelerate the information extraction of scientific literature.

# Introduction

Materials language processing (MLP) has emerged as a powerful tool in the realm of materials science research that aims to facilitate the extraction of valuable information from the scientific literature and the construction of a knowledge base [1, 2]. MLP leverages natural language processing (NLP) techniques to analyse and understand the language used in materials science texts, enabling the identification of key materials and properties and their relationships [3-6]. Despite significant advancements in MLP, challenges remain that hinder its practical applicability and performance. One key challenge lies in the availability of labelled datasets for training machine-learning models, as creating such datasets can be time-consuming and labour-intensive [4]. Additionally, fine-tuning pretrained large language models (LLMs) for knowledge-intensive MLP requires a large amount of training data to achieve satisfactory performance, limiting their effectiveness in scenarios with limited labelled data.

In this study, we propose a pipeline that employs the power of a generative pretrained transformer (GPT) [7] for solving MLP tasks. GPT is a state-of-the-art LLM that has demonstrated remarkable performance in various NLP tasks, such as text generation, translation, and comprehension. We aim to address the limitations of existing models and improve the practical applicability and performance of knowledge-intensive MLP tasks by employing the generative model. Our study focuses on three key MLP tasks: text classification, named entity recognition (NER), and extractive question answering (QA).

First, we present a document-classification method that leverages the strengths of zero-shot (without training data) and few-shot (with few training data) learning models, which show promising performance even with limited training data. This approach demonstrates the potential to achieve high accuracy in filtering relevant documents without fine-tuning based on a large-scale dataset. Furthermore, we propose an entity-centric prompt engineering method for NER, the performance of which surpasses that of previous fine-tuned models on multiple datasets. By carefully constructing prompts that guide the GPT model towards recognising and tagging materials-related entities, we enhance the accuracy and efficiency of entity recognition in materials science texts. Finally, we introduce a GPT-enabled extractive QA model that demonstrates improved performance in providing precise and informative answers to questions related to materials science. By fine-tuning the GPT model on materials-science-specific QA data, we enhance its ability to comprehend and extract relevant information from the scientific literature.

Through our experiments and evaluations, we validate the effectiveness of GPT-enabled MLP models, analysing their cost, reliability, and accuracy to advance materials science research. Furthermore, we discuss the implications of GPT-enabled models for practical tasks, such as entity tagging and annotation evaluation, shedding light on the efficacy and practicality of this approach. In summary, our research presents a significant advancement in MLP through the integration of GPT models. By leveraging the capabilities of GPT, we aim to overcome limitations in its practical applicability and performance, opening new avenues for extracting knowledge from materials science literature.

## Results

**Workflow of GPT-enabled materials language processing pipeline**

Fig. 1 presents an overview of our GPT-enabled MLP pipeline, which uses the embedding module and prompt–completion module of GPT-series models for text classification, NER, and extractive QA.

**Text classification**

Text classification, a fundamental task in NLP, involves categorising textual data into predefined classes or categories[8] (MLP task descriptions in Supporting Information). Text classification in materials science has been actively used for filtering valid documents from the retrieval results of search engines or identifying paragraphs containing information of interest [6, 9, 10]. For example, some researchers have attempted to classify the abstracts of battery-related papers from the results of searching with keywords such as '*battery*' or '*battery materials*', which is the starting point of extracting battery-device information from the literature [11]. Furthermore, paragraph-level classification models have been developed to find paragraphs of interest using a statistical model such as Latent Dirichlet allocation or machine-learning models such as random forest or BERT classifier [9, 12, 13], e.g., for solid-state synthesis, gold-nanoparticle synthesis, multiclass of solution synthesis.

**Battery-materials-related paper classification.** Some researchers have attempted to construct a battery database using NLP techniques applied to research papers[11]. The authors reported a dataset specifically designed for categorising papers relevant to battery research. Specifically, there are 46,663 labelled datasets for which the labels are battery or non-battery are publicly available (Fig S1.a). They annotated a substantial amount of data and created a classification model with complex structure by using multiple BERT-based models. Despite the reported SOTA performance is an accuracy of 97.5%, precision of 96.6%, and recall of 99.5%, such models require extensive training data and complex structures, and thus, we attempted to develop a simple, GPT-enabled model that can achieve high performance using only a small dataset. Specifically, we tested zero-shot learning and few-shot learning models based on GPT 3.5 for this classification task.

Zero-shot learning with embedding [14] allows models to make predictions or perform tasks without fine-tuning with human-labelled data. The zero-shot model works based on the embedding value of a given text, which is provided by GPT embedding modules. Using the distance between a given paragraph and predefined labels in the embedding space, which numerically represent their semantic similarity, paragraphs are classified with labels. For example, if one uses the model to classify an unseen text with the label of either "*batteries*" or "*solar cells*", the model will calculate the distance between the embedding value of the text and that of 'batteries' or 'solar cells', selecting the label with higher similarity in the embedding space.

Below are the results of the zero-shot text classification model using the text-embedding-ada-002 model of GPT (Fig 2.a). First, we tested the original label pair of the dataset [11], that is, "*battery*" vs. "*non-battery*" ('original labels' of Fig 2.a). The performance of the existing label-based model was low, with an accuracy and precision of 63.2%, because the difference between the embedding value of "*battery*" and that of "*non-battery*" was small, indicating that the model judges the two labels to be semantically similar. The model tended to predict most of the observations to have positive labels (recall ≈100%), as most of the papers in the dataset were collected through a keyword-based search, directly referring to the word "battery". Considering that the True label should indicate battery-related papers and the False label would result in the complementary dataset, we designed the label pair as "*battery materials*" vs. "*diverse domains*" ('crude labels' of Fig 2.a). We successfully improved the performance, achieving an accuracy of 87.3%, precision of 84.5%, and recall of 97.9%, by specifying the meaning of the false label.

To further reduce the number of false positives, we designed the labels in an explicit manner, i.e., "*battery materials*" vs. "*medical and psychological research*" ('designated labels' of Fig 2.a). Here, the false label was selected from the results of randomly sampled papers from the non-battery set. Interestingly, we obtained slightly improved performance (accuracy, recall, and precision of 91.0%, 88.6%, and 98.3%). We were able to achieve even higher performance (ACC: 93.0, PRE: 90.8, REC: 98.9) if the labels were made even more verbose: "*papers related to battery energy materials*" vs. "*medical and psychological research*" ('verbose labels' of Fig 2.a). Although these values are relatively lower than those of the SOTA model, it is noteworthy that acceptable text-classification performance was achieved without exhaustive human labelling, as the proposed model is based on zero-shot learning with embeddings. To summarize, we confirmed that selecting labels that well represent the valid paper set ("*battery*") and the complement ("*non-battery*") is a key determinant in improving the performance of zero-shot learning. These results imply that classifying a specific set among the paper data set in materials science can be achieved without labelling with zero-shot methods if a proper label corresponding to a representative embedding value for each category is selected.

Next, the improved performance of few-shot text classification models is demonstrated in Fig 2.b. In few-shot learning models, we provide the selected number of labelled data to the model. We tested 2-way 1-shot and 2-way 5-shot models, which means that there are two labels and one/five labelled data for each label are granted to the models. The example prompt is given in Fig 2.c. The 2-way 1-shot models resulted in an accuracy of 95.7%, which indicates that providing just one example for each category has a significant effect on the prediction. Furthermore, increasing the number of examples leads to improved performance, where the accuracy, precision, and recall are 96.1%, 95.0%, and 99.1%. Finally, we used the fine-tuning module of the GPT-3 davinci model with 1,000 prompt–completion examples. The fine-tuning model performs a general binary classification of texts by learning the examples while no longer using the embeddings of the labels, in contrast to few-shot learning. In our test, the fine-tuning model yielded high performance, that is, an accuracy of 96.6%, precision of 95.8%, and recall of 98.9%, which are close to those of the SOTA model.

Here, we emphasise that the GPT-enabled models can achieve acceptable performance even with the small number of datasets, although they slightly underperformed the BERT-based model trained with a large dataset.

In addition to the accuracy, we investigated the reliability of our GPT-based models and the SOTA models in terms of calibration. The reliability can be evaluated by measuring the expected calibration error (ECE) score [15]. The ECE score assesses the calibration of probabilistic predictions of models and is calculated as follows:

$$\text{ECE} = \sum_{m=1}^{M} \frac{|B_m|}{n} |acc(B_m) - conf(B_m)|,$$

where the dataset is divided into M interval bins based on confidence, and $B_m$ is the set of indices of samples of which the confidence scores fall into each interval, while $acc(B_m)$ and $conf(B_m)$ are the average accuracy and confidence for each bin, respectively.

A lower ECE score indicates that the model's predictions are closer to being well-calibrated, ensuring that the confidence of a model in its prediction is similar to the actual accuracy of the model [16]. The log probabilities of GPT-enabled models were used to compare the accuracy and confidence. The ECE score of the SOTA (batteryBERT-cased) model is 0.03, whereas those of the 2-way 1-shot model, 2-way 5-shot model, and fine-tuned model were 0.05, 0.07, and 0.07, respectively. Considering a well-calibrated model typically exhibits an ECE of less than 0.1, we conclude that our GPT-enabled text classification models provide high performance in terms of both accuracy and reliability with less cost. The lowest ECE score of the SOTA model shows that the BERT classifier fine-tuned for the given task was well-trained and not overconfident, potentially owing to the large and unbiased training set. The GPT-enabled models also show acceptable reliability scores, which is encouraging when considering the amount of training data or training costs required. In summary, we expect the GPT-enabled text-classification models to be valuable tools for materials scientists with less machine-learning knowledge while providing high accuracy and reliability comparable to BERT-based fine-tuned models.

**Named entity recognition**

NER is one of the representative NLP techniques for information extraction [17]. NER aims to identify and classify named entities within text (MLP task descriptions in Supporting Information). Here, named entities refer to real-world objects such as persons, organisations, locations, dates, and quantities [18]. In materials science, many researchers have developed NER models for extracting structured summary-level data from unstructured text. For example, domain-specific pretrained language models such as SciBERT [19], MatBERT [5], MatSciBERT [2], and MaterialsBERT [20] were used to extract specialised information from materials science literature, thereby extracting entities on solid-state materials, doping, gold nanoparticles (AuNPs), polymers, electrocatalytic $CO_2$ reduction, and solid oxide fuel cells from a large number of papers [5, 6, 20, 21]. In this work, we used the three publicly available datasets, which include human-labelled entities on solid-state materials, doped materials, and AuNPs, to compare the performance of our GPT-enabled models and prior ones.

**Solid-state materials entity recognition.** The solid-state materials dataset includes annotations on the following categories: inorganic materials (MAT), symmetry/phase labels (SPL), sample descriptors (DSC), material properties (PRO), material applications (APL), synthesis methods (SMT), and characterisation methods (CMT) [21]. For example, MAT indicates inorganics solid/alloy materials or non-gaseous elements such as '*BaTiO$_3$*,' '*titania,*' or '*Fe*'. SPL indicates the name for crystal structures and phases such as '*tetragonal*' or a symmetry label such as '*Pbnm*' (Fig. S1.b).

Because the fine-tuning model requires prompt–completion examples as a training set, the NER datasets are pre-processed as follows: the annotations for each category are marked with the special tokens [22], and then, the raw text and marked text are used as the prompt and completion, respectively. For example, if the input text is "LiCoO2 and LiFePO4 are used as cathodes of secondary batteries", the prompt–completion pair can be generated. The prompt is the same as the input text, and the completion for each category is as follows:

MAT model → Completion: "*LiCoO2 and LiFePO4 are used as cathodes of secondary batteries*" / completion: "*@@LiCoO2## and @@LiFePO4## are used as cathodes of secondary batteries.*"

APL model → Completion: "*LiCoO2 and LiFePO4 are used as cathodes of secondary batteries*" / completion: "*LiCoO2 and LiFePO4 are used as @@cathodes of secondary batteries##.*"

One of the examples used in the training set is shown in Fig 3.d. After pre-processing, we tested fine-tuning modules of GPT-3 such as davinci models. The performance of our GPT-enabled NER models was compared with that of the SOTA model in terms of recall, precision, and F1 score. Fig 3.a shows that the GPT model exhibits a higher recall value in the categories of CMT, SMT, and SPL and a slightly lower value in the categories of DSC, MAT, and PRO compared to the SOTA model. However, for the F1 score, our GPT-based model outperforms the SOTA model for all categories because of the superior precision of the GPT-enabled model (Fig. 3b–c). The high precision of the GPT-enabled model can be attributed to the generative nature of GPT models, which allows coherent and contextually appropriate output to be generated.

**Doped materials recognition.** The doped materials entity dataset [5] annotates the base material (BASEMAT), the doping agent (DOPANT), and quantities associated with the doped material such as the doping density or the charge carrier density (DOPMODQ), with specific examples provided in Fig. S1.b. The SOTA model for this dataset had F1 scores of 72, 82, and 62 for BASEMANT, DOPANT, and DOPMODQ, respectively. We analysed this dataset using fine-tuning modules of GPT-3 such as the davinci model. The prompt–completion sets were constructed similarly to the previous NER task. As reported in Fig 4.a, the fine-tuning GPT-3 davinci model showed high precision of 93.4, 95.6, and 92.7 for the three categories, BASEMAT, DOPANT, and DOPMODQ, respectively, while yielding relatively lower recall of 62.0, 64.4, and 59.4, respectively (Fig 4.a). These results imply that the doped materials entity dataset may have diverse entities for each category but that there is not enough data for training to cover the diversity. In addition, the GPT-based model's F1 scores of 74.6, 77.0, and 72.4 surpassed or closely approached those of the SOTA model (matBERT-uncased), which were recorded as 72, 82, and 62, respectively (Fig 4.b).

**AuNPs entity recognition.** The AuNPs entity dataset annotates the descriptive entities (DES) and the morphological entities (MOR) [12], where DES includes '*dumbbell-like*' or '*spherical*' and MOR includes noun phrases such as '*nanoparticles*' or '*AuNRs*'. More specific examples are provided in Fig. S1.c. The SOTA model for this dataset is reported as the matBERT-based model whose F1 scores for DES and MOR are 0.67 and 0.92, respectively[5].

We used the few-shot learning [23] of the GPT-3.5 model (text-davinci-003) for the AuNPs entities dataset. Similar to the previous NER task, we designed the prompt to randomly select the three ground-truth examples (pair of text and the text with named entities) from the training set when extracting the named entities from the given text in the test set (random retrieval). These simple methods yield high recall performance of 63% and 97% for the DES and MOR categories, respectively. Here, it is noteworthy that prompts with the ground-truth examples can provide improved results on DES and MOR entity recognition, considering the recall values of 52% and 64% reported in prior works [12] (Fig. S2). However, the F1 score of this few-shot learning model was lower than that of the SOTA model ('random retrieval' of Fig 4.c). Furthermore, we tested the effect of adding a phrase that directly specifies the task to the existing prompt; e.g., "*The task is to extract the descriptive entities of materials in the given text*" ('task-informed random retrieval' of Fig 4.c). The example prompt is shown in Fig 4.d. Some performance improvements, namely a 1%–2% increase in recall and a 6%–11% increase in precision, were observed.

Finally, to more elaborately perform the few-shot learning, "similar" ground-truth examples to each test set, that is, the examples for which the document embedding value are similar to that of each test set, were selected for the NER extraction in the test set ('kNN retrieval' of Fig 4.c). Interestingly, compared to the performance of the previous method (i.e., task-informed random retrieval), we confirmed that the recall value of the kNN method was the same or slightly lower and that the precision increased by 15%–20%. Particularly, the recall of DES was relatively low compared to its precision, which indicates that providing similar ground-truth examples enables more tight recognition of DES entities. In addition, the recall of MOR is relatively higher than the precision, implying that giving k-nearest examples results in the recognition of more permissive MOR entities (Fig S2). In summary, we confirmed the potential of the few-shot NER model through GPT prompt engineering and found that providing similar examples rather than randomly sampled examples and informing tasks had a significant effect on performance improvement. In terms of the F1 score, few-shot learning with the GPT-3.5 (text-davinci-003) model results in comparable MOR entity recognition performance as that of the SOTA model and improved DES recognition performance (Fig 4.c).

**Extractive question answering**

Extractive QA is a type of QA system that retrieves answers directly from a given passage of text rather than generating answers based on external knowledge or language understanding [24] (MLP task descriptions in Supporting Information). In materials science, the extractive QA task has received less attention as its purpose is similar to the NER task for information extraction, although battery-device-related QA models have been proposed [11]. Nevertheless, by enabling accurate information retrieval, advancing research in the field, enhancing search engines, and contributing to various domains within materials science, extractive QA holds the potential for significant impact.

**Battery-device-related question answering.** This dataset [11] consists of questions, contexts, and answers, and the questions are related to the principal components of battery systems, i.e., "*What is the anode?*", "*What is the cathode?*", and "*What is the electrolyte?*". The publicly available dataset includes 427 annotations, although the authors stated they made 272 manually labelled annotations. Unfortunately, we found redundant or incorrect annotations, e.g., when there is no mention of the anode in the given context, the question is about the anode and the answer is about the cathode. In the end, we refined the given dataset into 331 QA data.

Next, we reproduced the results of prior QA models including the SOTA model, BatteryBERT (cased), to compare the performances between our GPT-enabled models and prior models with the same measure. The performances of the models were newly evaluated with the average values of token-level precision and recall, which are usually used in QA model evaluation. In this way, the prior models were re-evaluated, and the SOTA model turned out to be batteryBERT (cased), identical to that reported (Fig 5.a).

We tested the zero-shot QA model using the text-davinci-003 model of GPT-3.5, yielding a precision of 60.92%, recall of 79.96%, and F1 score of 69.15%. These relatively low performance values can be derived from the domain-specific dataset, from which it is difficult for a vanilla model to find the answer from the given scientific literature text. Therefore, we added a task-informing phrase such as '*The task is to extract answers from the given text.*' to the existing prompt consisting of the question, context, and answer. Surprisingly, we observed an increase in performance, particularly in precision, which increased from 60.92% to 72.89%. By specifying that the task was to extract rather than generate answers, the accuracy of the answers appeared to increase. Next, we tested a fine-tuning module of GPT-3 models such as *davinci*. We achieved higher performance with an F1 score of 88.21% (compared to that of 74.48% for the SOTA model).

In addition to the improved performance, we were able to examine the possibility of correcting the existing annotations with our GPT-based models. As mentioned earlier, we modified and used the open QA data set. Here, in addition to removing duplicates or deleting unanswered data, finding data with incorrect answers was based on the results of the GPT model ([Fig 5.c](Fig 5.c)). For example, there is an incorrect question–answer pair: the anode materials are not mentioned in the given context and '*nano-meshed*' is mentioned as the cathode material; however, the annotated question is 'what is the anode material?', and the corresponding answer is '*nano-meshed*'. For this case, most BERT-based models yield the answer '*nano-meshed*' similar to the annotation, whereas the GPT models provide the answer '*the anode is not mentioned in the given text*'. In addition, there were annotations that could increase the confusion of the model by making each question–answer pair for the answer in which the two tokens were combined by OR. For example, GPT models answered "*sulfur or air cathode*", but the original annotations annotate '*sulfur*' and '*air*' as different answers.

## Discussion

This work presents a GPT-enabled pipeline for MLP tasks, providing guidelines for text classification, NER, and extractive QA. Through an empirical study, we demonstrated the advantages and disadvantages of GPT models in MLP tasks compared to the prior fine-tuned models based on BERT (Concluding remarks in Supporting Information). We note the potential limitations and inherent characteristics of GPT-enabled MLP models, which materials scientists should consider when analysing literature using GPT models. First, considering that GPT series models are autoregressive and generative, the additional step of examining whether the results are faithful to the original text would be necessary in MLP tasks, particularly information-extraction tasks [25]. In contrast, general MLP models based on fine-tuned LLMs do not provide unexpected prediction values because they are classified into predefined categories through cross entropy function. Similarly, given that GPT is a closed model that does not disclose the training details and the response generated carries an encoded opinion, the results are likely to be overconfident and influenced by the biases in the given training data [26]. Therefore, it is necessary to evaluate the reliability as well as accuracy of the results when using GPT-guided results for the subsequent analysis. Finally, the GPT-enabled model would face challenges in more domain-specific, complex, and challenging tasks (e.g., relation extraction, event detection, and event extraction) than those presented in this study, as it is difficult to explain the tasks in the prompt. For example, extracting the relations of entities would be challenging as it is necessary to explain well the complicated patterns or relationships as text, which are inferred through black-box models in general NLP models [27]. Nonetheless, GPT models will be effective MLP tools by allowing material scientists to more easily analyse literature effectively without knowledge of the complex architecture of existing NLP models. As LLM technologies advance, creating quality prompts that consist of specific and clear task descriptions, appropriate input text for the task, and consistently labelled results (i.e., classification categories) will become more important for materials scientists.

## Methods

**Prompt engineering**

We used the python library *openai* to implement the GPT-enabled MLP pipeline. We mainly used the prompt–completion module of GPT models for training examples for text classification, NER, or extractive QA. Given a sufficient dataset of prompt–completion pairs, a fine-tuning module of GPT-3 models such as *davinci* or *curie* can be used. The prompt–completion pairs are lists of independent and identically distributed training examples concatenated together with one test input. Otherwise, for few-shot learning, which makes the prompt consisting of the task-informing phrase, several examples and the input of interest, can be alternatives. Here, which examples to provide is important in designing effective few-shot learning. Similar examples can be obtained by calculating the cosine similarity between the training set for each test set.

Regarding the preparation of prompt–completion examples, we suggest some guidelines. Suffix characters in the prompt such as ' →' are required to clarify to the fine-tuned model where the completion should begin. In addition, suffix characters in the prompt such as ' \n\n###\n\n' are required to specify the end of the prediction. This is important when a trained model decides on the end of its prediction for a given input, given that GPT is one of the autoregressive models that continuously predicts the following text from the preceding text. That is, in prediction, the same suffix should be placed at the end of the input. In addition, prefix characters are usually unnecessary as the prompt and completion are distinguished. Rather than using the prefix characters, simply starting the completion with a whitespace character would produce better results due to the tokenisation of GPT models. In addition, this method can be economical as it reduces the number of unnecessary tokens in the GPT model, where fees are charged based on the number of tokens. We note that the maximum number of tokens in a single prompt–completion is 4097, and thus, counting tokens is important for effective prompt engineering; e.g., we used the python library *titoken* to test the tokenizer of GPT series models.

**GPT model usage guidelines**

After pre-processing, the splitting process of train, validation, and test set proceeds, and the dataset is divided by using the random seed and ratio used in previous studies. In the fine-tuning of GPT models, there are some hyperparameters such as the base model, batch size, number of epochs, learning rate multiplier, and prompt loss weight. The base models for which fine-tuning is available are GPT-3 models such as '*ada*', '*babbage*', '*curie*', and '*davinci*', which can be tested using the web service provided by OpenAI (https://gpttools.com/comparisontool). For a simple prompt–completion task such as zero-shot learning and few-shot learning, GPT-3.5 models such as '*text-davinci-003*' can be used. The batch size can be dynamically configured and its maximum is 256; however, we recommend 1% or 0.2% of the training set. The learning rate multiplier adjusts the models' weights during training, and a high learning rate leads to a sub-optimal solution, whereas a low one causes the model to converge too slowly or find a local minimum. The default values are 0.05–0.2 depending on the batch size, and we set the learning rate multiplier as 0.01. The prompt loss weight is the weight to use for loss on the prompt tokens, which should be reduced when prompts are relatively long to the corresponding completions to avoid giving undue priority to prompt learning over the completion learning. We set the prompt loss weight as 0.01.

With the fine-tuned GPT models, we can infer the completion for a given unseen dataset that ends with the pre-defined suffix. Here, some parameters such as the temperature, maximum number of tokens, and top P can be determined according to the purpose of analysis. First, temperature determines the randomness of the completion generated by the model, ranging from 0 to 1. For example, higher temperature leads to more randomness in the generated output, which can be useful for exploring creative or new completions (e.g., generative QA). In addition, lower temperature leads to more focused and deterministic generations, which is appropriate to obtain more common and probable results, potentially sacrificing novelty. We set the temperature as 0, as our MLP tasks concern the extraction of information rather than the creation of new tokens. The maximum number of tokens determines how many tokens to generate in the completion. If the ideal completion is longer than the maximum number, the completion result may be truncated; thus, we recommend setting this hyperparameter to the maximum number of tokens of completions in the training set (e.g., 256 in our cases). In practice, the reason the GPT model stops producing results is ideally because a suffix has been found; however, it could be that the maximum length is exceeded. The top P is a hyperparameter about the top-p sampling, i.e., nucleus sampling, where the model selects the next word based on the most likely candidates, limited to a dynamic subset determined by a probability threshold ($p$). This parameter promotes diversity in generated text while allowing control over randomness.

**Performance evaluation**

We evaluated the performance of text classification, NER, and QA models using different measures. The fine-tuning module provides the results of accuracy, actually the exact-matching accuracy. Therefore, post-processing of the prediction results was required to compare the performance of our GPT-based models and the reported SOTA models. For the text classification, the predictions refer to one of the pre-defined categories. By comparing the category mentioned in each prediction and the ground truth, the accuracy, precision, and recall can be measured. For the NER, the performance such as the precision and recall can be measured by comparing the index of ground-truth entities and predicted entities. Here, the performance can be evaluated strictly by using an exact-matching method, where both the start index and end index of the ground-truth answer and prediction result match. The boundaries of named entities are likely to be subjective or ambiguous in practice, and thus, we recommend the boundary-relaxation method to generously evaluate the performance, where a case that either the start or end index is correct is considered as a true positive [28]. For the extractive QA, the performance is evaluated by measuring the precision and recall for each answer at the token level and averaging them. Similar to the NER performance, the answers are evaluated by measuring the number of tokens overlapping the actual correct answers.

# Acknowledgements

**Funding**

This work was supported by the National Research Foundation of Korea funded by the Ministry of Science and ICT (NRF-2021M3A7C2089739) and Institutional Projects at the Korea Institute of Science and Technology (2E31742 and 2E32533).

**Author contributions**

Jaewoong Choi: Conceptualisation, Methodology, Programming, Data analysis, Visualisation, Interpretation, Writing – original draft, Writing – review & editing.

Byungju Lee: Conceptualisation, Interpretation, Writing – review & editing, Supervision, Resources, Funding acquisition.

**Competing interests**

The authors declare no competing interests.

**Data and materials availability**

Data, code and materials used in this study are available in https://github.com/AIHubForScience/GPT_MLP.

# Figures

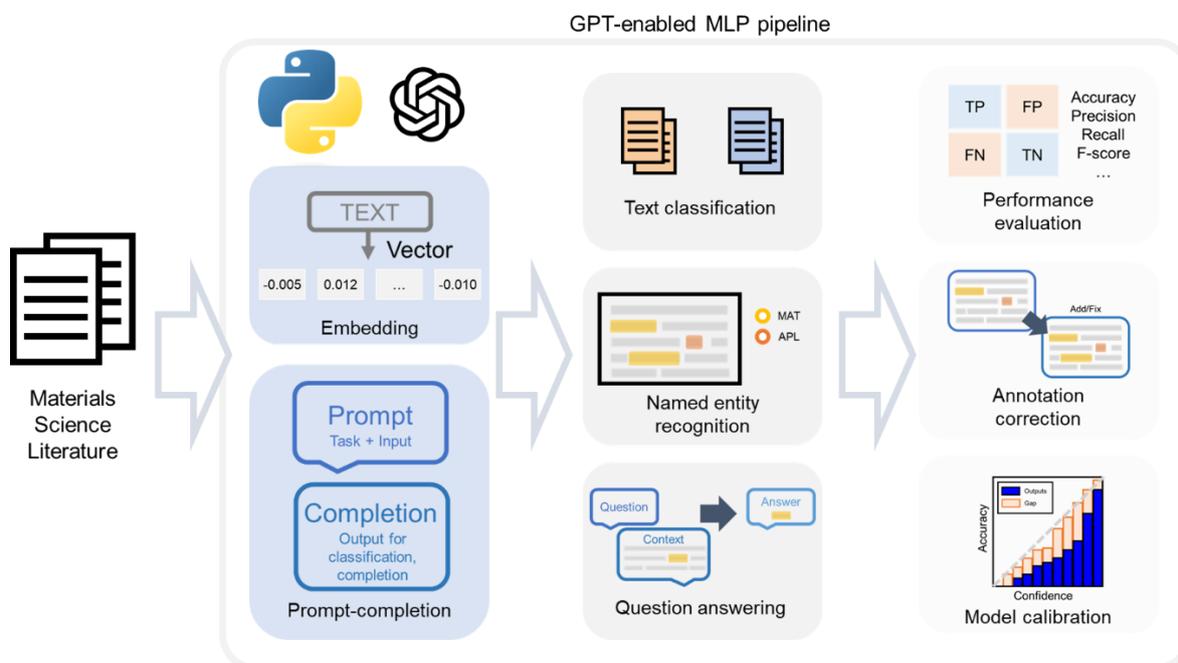

**Fig. 1. Workflow of GPT-enabled MLP pipeline**. In this work, we suggest the use of GPT-series models for three representative MLP tasks (text classification, NER, and QA) using publicly available datasets. The datasets, with a large number of human-labelled annotations, are used as ground-truth examples of MLP tasks, i.e., valid documents, named entities, and answers. We compare the performance of our GPT-based models with that of the state-of-the-art (SOTA) model, discuss the availability of our results as annotation correction, and investigate the reliability of the models in terms of calibration. Specifically, the proposed GPT-enabled MLP pipeline analyses text information of materials science literature using the embedding module and prompt–completion module of GPT series models.

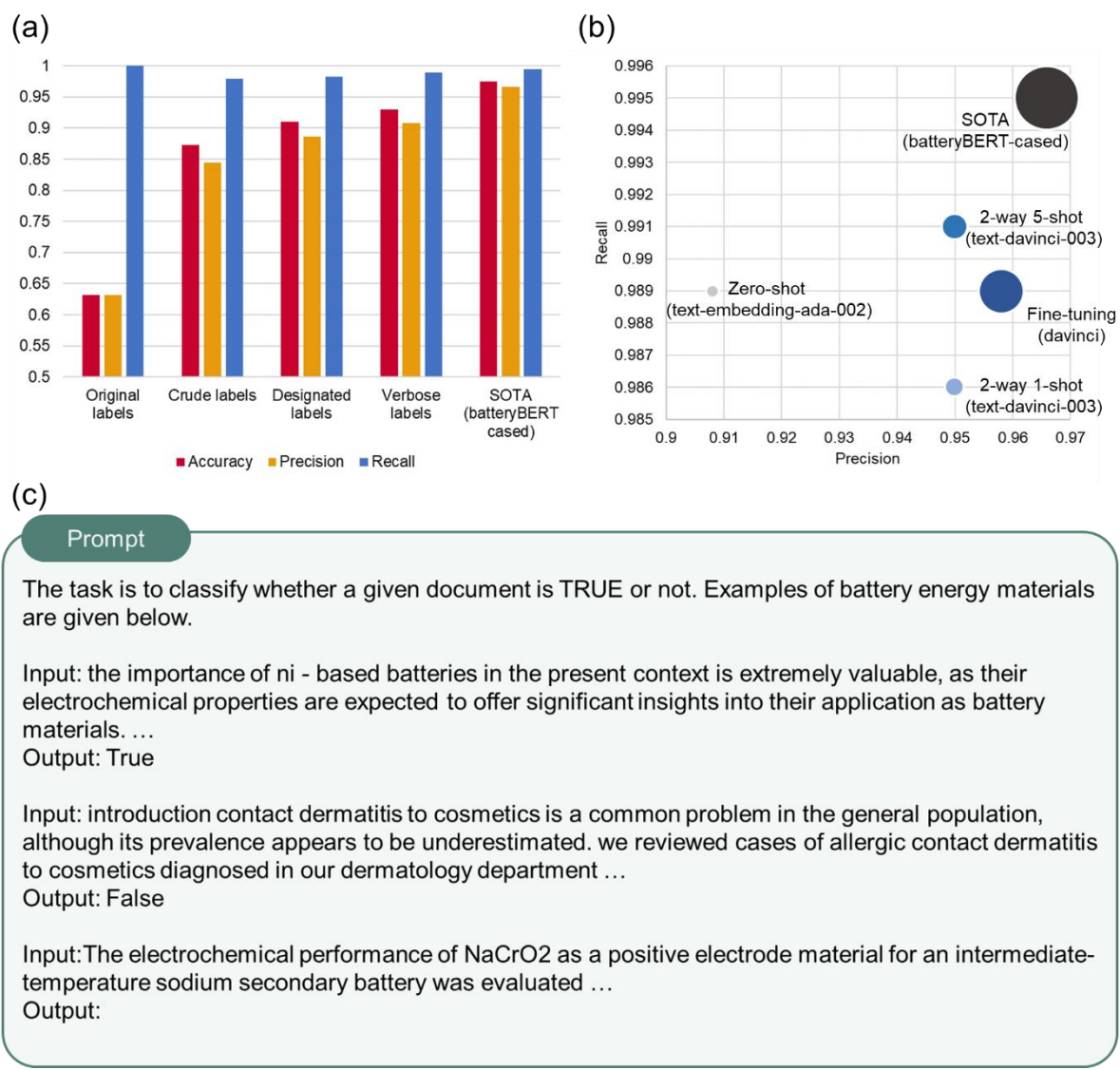

**Fig. 2. Results of GPT-enabled text classification models.** (**A**) Results of zero-shot learning with GPT embedding. The accuracy, precision, and recall are reported. (**B**) Comparison of zero-shot learning, few-shot learning, and fine-tuning results. The horizontal and vertical axes are the precision and recall of each model, respectively. The node colour and size are based on the rank of accuracy and the dataset size, respectively. (**C**) Example of prompt engineering for 2-way 1-shot learning, where the task description, one example for each category, and input abstract are given.

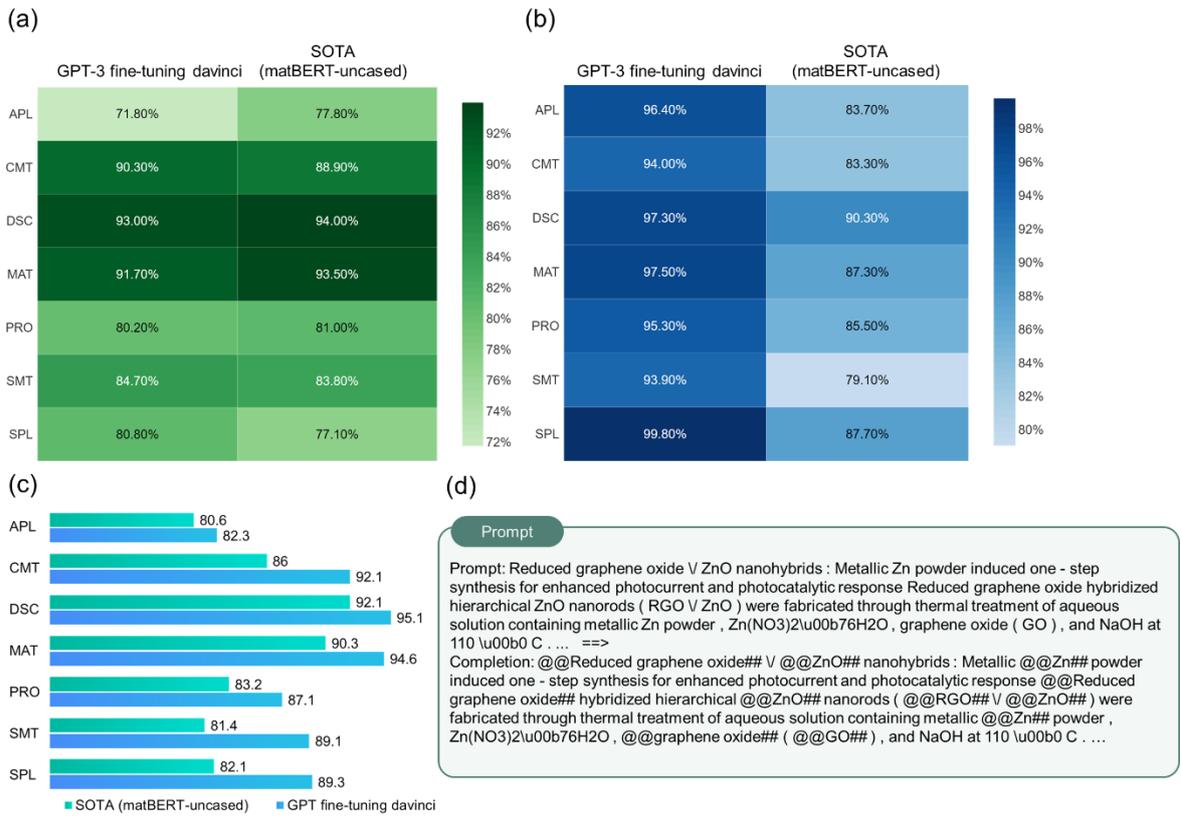

**Fig. 3. Performance of GPT-enabled NER models on solid-state materials compared to the SOTA model (matBERT-uncased).** The proposed models are based on fine-tuning modules based on prompt–completion examples. **(A–C)** Comparison of recall, precision, and F1 score between our GPT-enabled model and the SOTA model for each category. **(D)** Example of prompt–completion for MAT entity recognition.

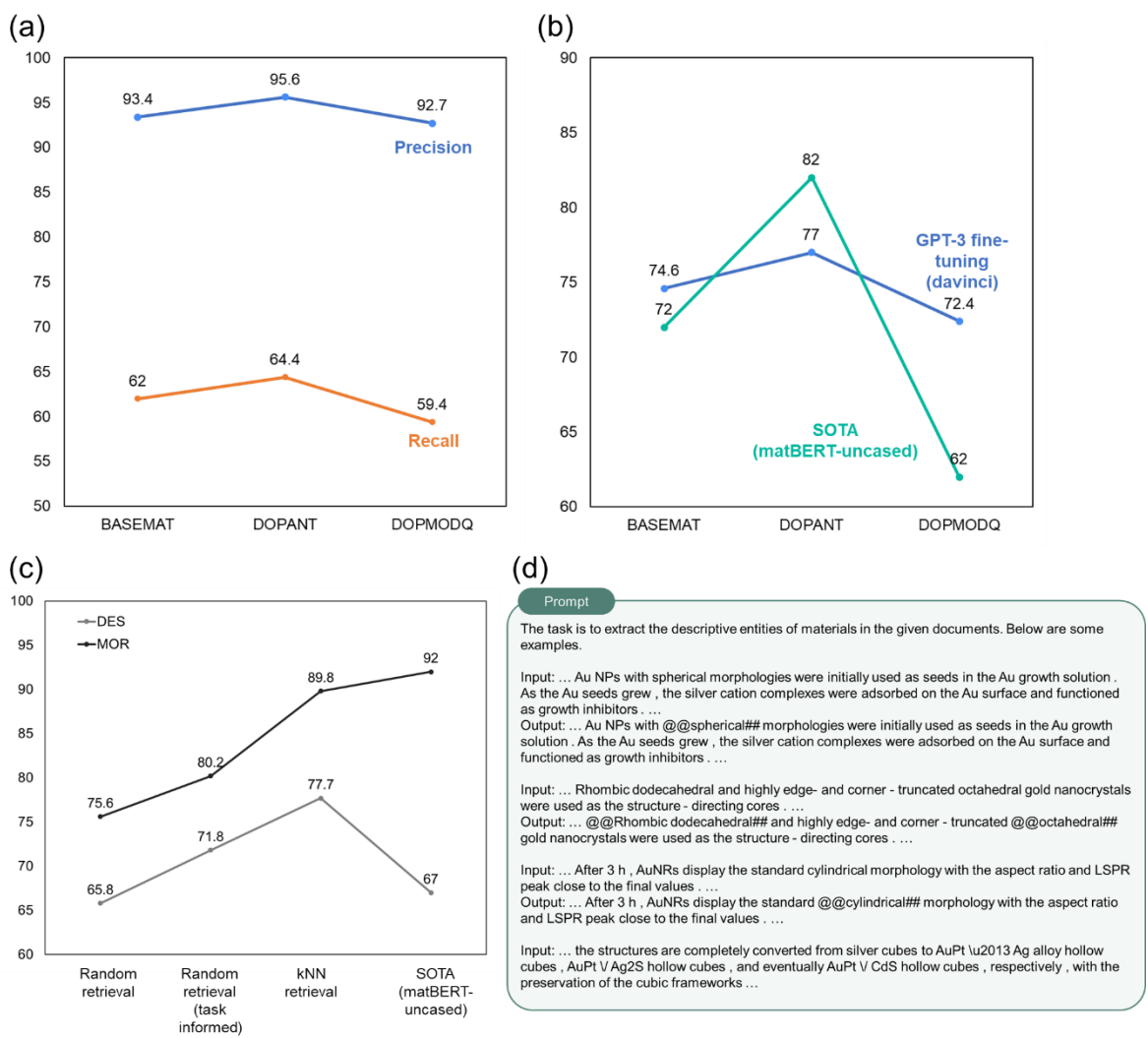

**Fig. 4. Performance of GPT-enabled NER models on doped materials and AuNPs, compared to the SOTA model.** (**A**) Doped materials entity recognition performance of fine-tuning of GPT 3 (davinci), (**B**) doped materials entity recognition performance (F1 score) comparison between SOTA (matBERT-uncased) and fine-tuning of GPT 3 davinci, (**C**) AuNPs entity recognition performance (F1-score) comparisons between GPT 3.5 davinci (random retrieval, task-informed random retrieval, kNN retrieval) and SOTA (matBERT-uncased) model, (**D**) Example of prompt for DES entity recognition (task informed random retrieval).

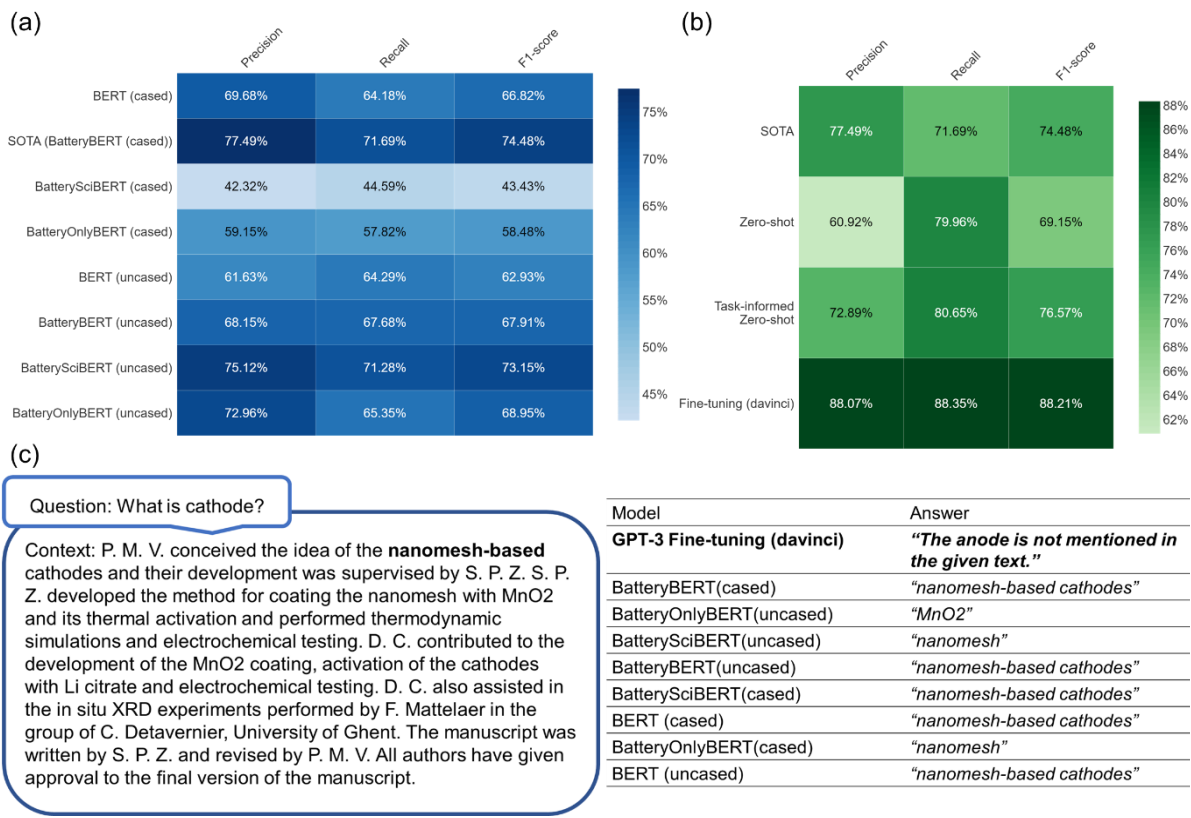

**Fig. 5. Performance of GPT-enabled QA model. (A)** Reproduced results of BERT-based model performances, **(B)** comparison between the SOTA and GPT-3 fine-tuning model (davinci), **(C)** correction of wrong annotations in QA dataset, and prediction result comparison of each model. Here, the difference in the cased/uncased version of the BERT series model is the processing of capitalisation of tokens or accent markers, which influenced the size of vocabulary, pre-processing, and training cost.